\documentclass[journal]{IEEEtran}
\usepackage{cite}
\usepackage{amsmath,amssymb,amsfonts}
\usepackage{graphicx}
\usepackage{textcomp}
\usepackage{booktabs}
\usepackage{multirow}
\usepackage{array}
\usepackage{tabularx}
\usepackage{xspace}
\usepackage{float}
\usepackage{url}

\def\BibTeX{{\rm B\kern-.05em{\sc i\kern-.025em b}\kern-.08em
    T\kern-.1667em\lower.7ex\hbox{E}\kern-.125emX}}

\newcommand{\SAC}{\text{SAC}\xspace}

\begin{document}

\title{Spatial Artifact Coherence Determines Codec Robustness in Patch-Based rPPG}

\author{Achraf~Ben~Ahmed,~PlesmoSense~SARL\\achraf@plesmosense.com%
\thanks{All experimental results, figures, and algorithm implementations presented in this manuscript are fully reproducible. The source code is publicly available at \texttt{github.com/baachraf/patchpca-codec-rppg}.}}

\maketitle

\begin{abstract}
Remote photoplethysmography (rPPG) achieves low heart-rate error on uncompressed benchmarks yet is deployed over compressed video channels in telehealth, neonatal ICU, and driver fatigue applications. No prior work identifies the physical quantity determining when spatial decomposition outperforms global-projection methods under codec compression.
We propose Spatial Artifact Coherence (SAC), defined as the ratio of off-diagonal to diagonal energy in the $4{\times}4$ inter-patch Green-channel covariance matrix (bandpass 0.75--2.5\,Hz), and the PatchPCA algorithm family (four codec-aware rPPG algorithms). We evaluate 280 subjects across three public datasets, 11 codec degradation variants (MPEG-4, H.265, H.264, JPEG, chroma subsampling), and 13 algorithms via Wilcoxon tests (BH-FDR, $q < 0.05$, 904 tests).
SAC explains 93.8\% of between-variant variance in PCA advantage ($r = +0.969$), with zero overlap between codec families: non-MPEG-4 variants cluster at SAC 0.10--0.18 with 84--90\% PCA win rates, while MPEG-4 variants cluster at SAC 0.48--0.59 with 61\% win rate and a 5.8$\times$ reduction in mean improvement. Within subjects, 78\% confirm the expected pattern ($p < 10^{-22}$, $d_z = 0.73$). Within-variant subject-level SAC correlation is $r = +0.099$, confirming SAC classifies codec families rather than predicting individual outcomes. MPEG-4's effect is structural (macroblock DCT geometry, not noise amplitude), governed by source codec state, not resolution.
P-Hybrid is identified as the most deployment-robust algorithm. Two necessary operating conditions for PatchPCA advantage are established: SAC~$< 0.30$ and low-to-moderate motion, directly ruling out raw-to-MPEG-4 transcoding pipelines.
SAC provides a physically grounded metric for codec-aware rPPG algorithm selection in clinical remote monitoring systems.
\end{abstract}


\section{Introduction}
\label{sec:intro}

\IEEEPARstart{C}{ompressed} video is the universal delivery medium for clinical remote monitoring. H.264 is the dominant codec for streamed internet video; clinical IP cameras deployed in neonatal intensive care units, telehealth platforms, and automotive fatigue detection systems encode at 85--1500\,kbps in MPEG-4 or H.264. Remote photoplethysmography (rPPG) extracts the cardiac pulse from facial colour fluctuations in ordinary video, enabling non-contact physiological sensing without electrodes or physical contact. The method is increasingly evaluated for clinical deployment in exactly these compressed-channel environments. Yet rPPG research assesses performance almost exclusively on uncompressed or near-lossless video: algorithms such as CHROM~\cite{b3} and POS~\cite{b4} achieve mean absolute errors below 2\,BPM on benchmarks such as UBFC-rPPG~\cite{b17}, which captures at roughly 218\,000\,kbps. Performance under the 85--1500\,kbps conditions that deployed systems process has not been characterised.

The rPPG literature divides into two paradigms. Global-projection methods (CHROM~\cite{b3}, POS~\cite{b4}) project whole-region RGB onto a skin-tone-orthogonal plane; they degrade monotonically with compression but do not fail catastrophically. Spatial decomposition methods apply PCA across multiple facial patches, relying on the assumption that codec artefacts are spatially incoherent. When this holds, PCA isolates the pulse as the dominant eigenvector; when it fails, the artefact captures the first eigenvector and the advantage reverses.

Whether codec artefacts satisfy the spatial incoherence assumption, and which codec families violate it, has not been investigated. Existing compression studies measure rPPG performance as a function of compression strength, treating codec type as a categorical label rather than a source of distinct spatial artefact geometry~\cite{b13,b19,b20,b21}. No work proposes a metric characterising the spatial coherence of codec artefacts across facial patches, nor derives operating conditions for PCA-based rPPG from such a metric. The present study addresses this gap directly.

\subsection{Motivating Observation}

MCD~\cite{b16} captures the same subjects with three cameras at identical 640$\times$480 resolution. PCA variants outperform CHROM by 3.29\,BPM on IriunWebcam ($\sim$792\,kbps) but only 1.75\,BPM on FullHDwebcam ($\sim$1346\,kbps): the relative algorithm ordering changes with acquisition codec despite matched subjects, session, and resolution. This cannot be explained by resolution, bitrate, or signal quality; it points to a structural codec property acting on inter-patch covariance.

\subsection{Related Work}

\subsubsection{Foundational rPPG methods and global-projection baselines}
The origins of camera-based pulse extraction lie in non-contact photoplethysmography using green-channel reflectance~\cite{b1}, subsequently extended to multi-channel blind source separation via ICA applied to raw RGB frames~\cite{b2}. The two algorithms that define the global-projection paradigm are CHROM~\cite{b3} and POS~\cite{b4}. CHROM decomposes RGB colour into a chrominance plane orthogonal to the specular reflection component; POS projects onto the plane orthogonal to the mean skin tone. Both operate on whole-ROI mean signals, preserving no inter-patch spatial structure. The spectral redundancy approach (2SR)~\cite{b5} achieves denoising by enforcing spectral consistency across spatial regions, but its reliance on coherent spectral peaks makes it susceptible to the structured inter-patch covariance noise that we identify as SAC.

\subsubsection{Multi-patch, spatial decomposition, and deep learning}
Prior work has exploited spatial facial diversity via PCA on head-motion landmarks~\cite{b6}, self-adaptive matrix completion~\cite{b7}, and spatial subspace rotation~\cite{b8}. Deep-learning methods including DeepPhys~\cite{b9}, RhythmNet~\cite{b10}, and PhysNet~\cite{b11} use end-to-end spatial-temporal networks; a recent survey covers more than thirty such pipelines~\cite{b12}. All of these approaches use spatial representations incompatible with the four-patch Green-channel covariance protocol employed here, and none characterises codec effects on inter-patch covariance. Deep-learning methods implicitly assume that training and deployment video share the same compression characteristics; a model trained on uncompressed video and deployed over MPEG-4 will encounter SAC patterns entirely absent from its training distribution. No published rPPG method reports per-codec-family performance statistics or provides a physically interpretable feature analogous to SAC.

\subsubsection{Video compression as a confound in rPPG}
McDuff \textit{et al.}~\cite{b13} showed H.264 CRF degrades rPPG SNR proportionally; Zhao \textit{et al.}~\cite{b14} proposed a deep pre-processing stage for partial signal recovery. Both treat codec as a scalar degradation parameter. Nowara, McDuff, and Veeraraghavan~\cite{b19} systematically evaluated H.264, H.265, and MPEG-4 across five CRF levels, identifying MPEG-4 as worst-performing — the most direct empirical precursor to this work — but used global-projection methods only and did not characterise the spatial covariance mechanism. Zhou \textit{et al.}~\cite{b20} compared seven codecs and recommended intra-frame H.265; Sporrer \textit{et al.}~\cite{b21} found performance collapse under high CRF for deep-learning rPPG. Wang \textit{et al.}~\cite{b22} proposed physiological-preserving compression; Comas \textit{et al.}~\cite{b23} proposed deep magnification for MPEG-4 video. No prior work reports the qualitative discontinuity between MPEG-4 and all other codec families, proposes a spatial coherence metric, or derives PCA-based operating conditions from such a metric.

\subsubsection{Eigenvalue and covariance-based blind source separation}
Parra and Sajda~\cite{b15} established generalised eigenvalue decomposition for blind source separation; rPPG covariance decomposition has been applied at the whole-frame level~\cite{b2,b6}. Our work uses the covariance matrix across spatially localised patches and employs it predictively: the off-diagonal-to-diagonal ratio (SAC) is applied to the raw input covariance before any algorithm runs to determine whether eigendecomposition will succeed under a given codec. No prior BSS or rPPG work uses such a ratio as a boundary condition for algorithm performance.

\textbf{Summary of the gap.} Across all five threads surveyed, no prior work (i) quantifies the spatial covariance structure of codec artefacts across multiple facial patches; (ii) identifies a metric linking codec geometry to algorithm boundary conditions; or (iii) demonstrates that source codec state, rather than resolution or bitrate, is the primary determinant of that structure.

\subsection{Contributions}

\begin{figure}[!t]
\centerline{\includegraphics[width=\columnwidth]{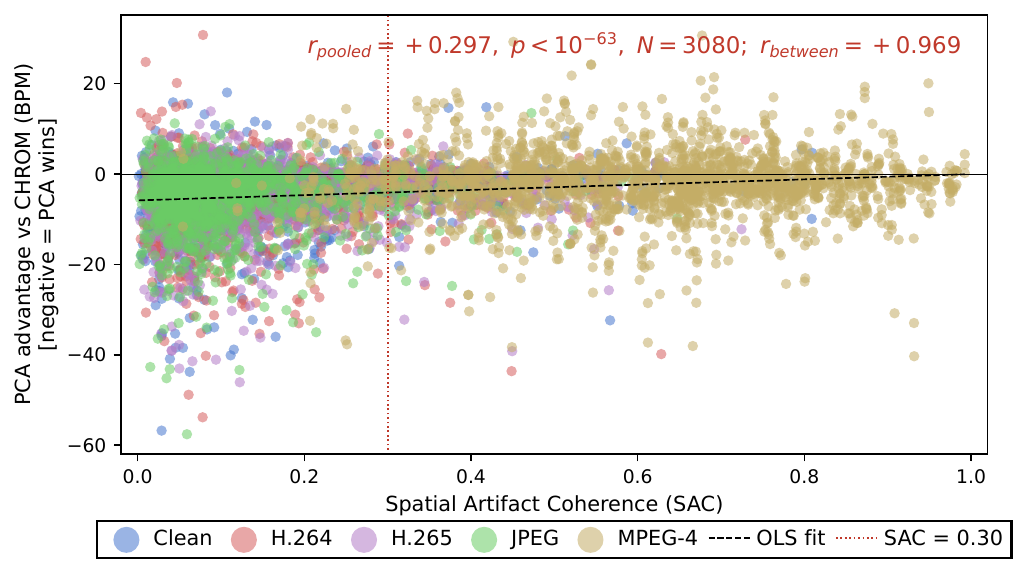}}
\caption{SAC versus $\Delta$MAE (PCA MAE $-$ CHROM MAE, BPM) across 3080 cells from 280 subjects. Non-MPEG-4 variants at SAC 0.10--0.18 (84--90\% PCA win rate); MPEG-4 at SAC 0.48--0.59 (61\% win rate). Between-variant $r = +0.969$ ($R^2 = 93.8\%$); within-variant $r = +0.099$. Within subjects, 78\% confirm the expected pattern ($p < 10^{-22}$, $d_z = 0.73$).}
\label{fig:scatter}
\end{figure}

\begin{figure}[!t]
\centerline{\includegraphics[width=\columnwidth]{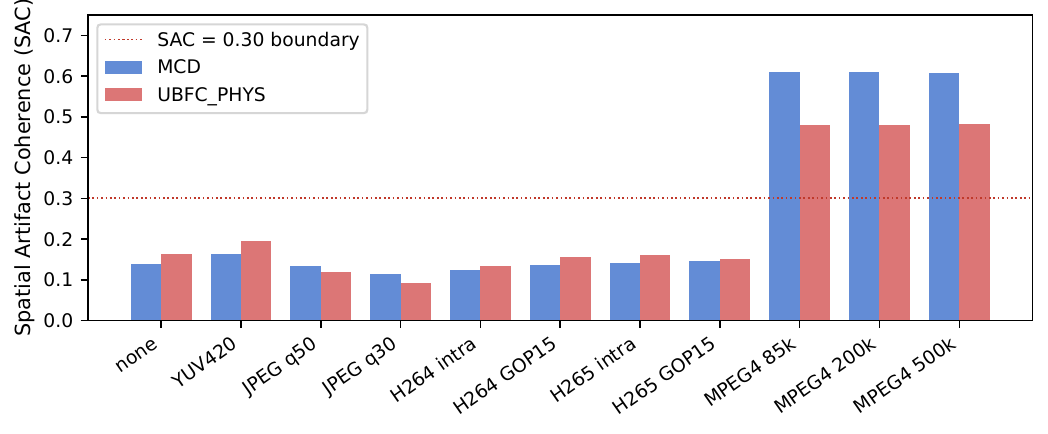}}
\caption{Off-diagonal to diagonal covariance energy ratio per degradation variant for MCD (left) and UBFC-PHYS (right). MPEG-4 uniquely elevates SAC at all three bitrates: $\sim$290$\times$ the clean baseline in MCD and $\sim$269$\times$ in UBFC-PHYS. H.265 GOP15 produces moderate elevation ($\sim$2.8--3.7$\times$), while all other variants are indistinguishable from clean ($\leq 1\times$). Increasing MPEG-4 bitrate reduces artefact amplitude but not spatial coherence; SAC is structural, not amplitude-dependent.}
\label{fig:sac_ratio}
\end{figure}

\begin{figure}[!t]
\centerline{\includegraphics[width=\columnwidth]{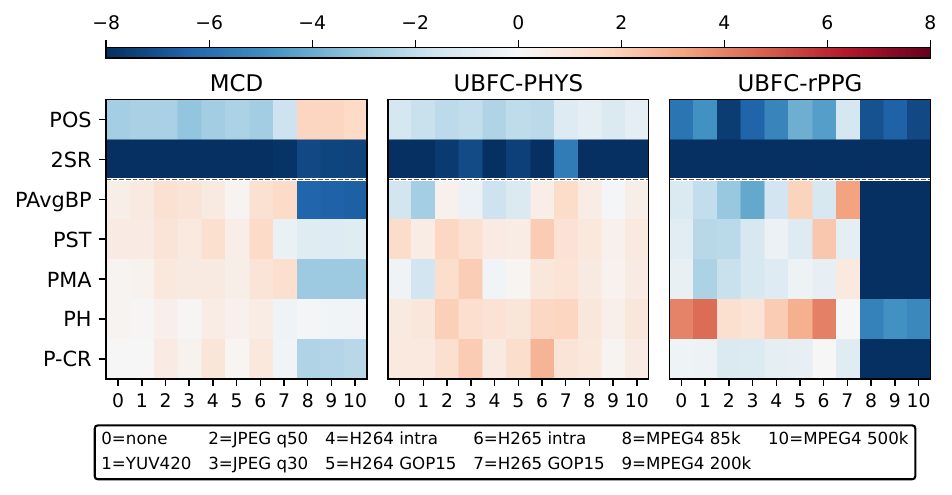}}
\caption{$\Delta$MAE heatmap (PCA $-$ CHROM, BPM) for five algorithms $\times$ 11 variants across MCD (left), UBFC-PHYS (centre), and UBFC-rPPG (right). Blue = PCA wins; red = CHROM wins. MPEG-4 columns reveal the source-codec-state effect: near-zero $\Delta$ on MCD (SAC = 0.59) versus strongly positive on UBFC-rPPG (SAC = 0.68). PAvgBP = PatchAvg\_Bandpass, PST = P-SpatialTemporal, PMA = P-Motion\_Adaptive, PH = P-Hybrid.}
\label{fig:heatmap}
\end{figure}

\begin{enumerate}
\item \textbf{Spatial Artifact Coherence (SAC).} We introduce SAC and validate it as the physical determinant of patch-PCA viability under codec compression. Across 280 subjects, three datasets, 11 degradation variants, and 13 algorithms: (i) SAC explains 93.8\% of between-variant variance in PCA advantage ($r = +0.969$ across 11 variants); (ii) the SAC distribution is bimodal with zero overlap: non-MPEG-4 at SAC 0.10--0.18, MPEG-4 at 0.48--0.59; (iii) PCA win rates drop from 84--90\% (non-MPEG-4) to 61\% (MPEG-4), with a 5.8$\times$ reduction in mean MAE improvement; (iv) within subjects, 78\% show the expected SAC--delta pattern ($p < 10^{-22}$, $d_z = 0.73$)~(Figs.~\ref{fig:scatter}~and~\ref{fig:sac_ratio}).

\item \textbf{PatchPCA algorithm family.} We propose four codec-aware rPPG algorithms: P-Hybrid (flagship, CHROM-anchored residual PCA), P-Motion\_Adaptive (motion-gated PCA, H.265 GOP15 specialist), P-SpatialTemporal (joint spatial-temporal decomposition, intra-codec specialist), and PatchAvg\_Bandpass (coherence-exploiting patch average, JPEG specialist). P-CodecRobust is included as an intentional negative design; its systematic failure on clean video establishes that codec-targeted corrections degrade performance when the target artefact is absent~(Figs.~\ref{fig:heatmap}~and~\ref{fig:profiles}).

\item \textbf{Deployment boundary conditions.} We empirically identify two necessary operating conditions for PatchPCA advantage: (i) SAC~$< 0.30$, ruling out all MPEG-4 pipelines operating on raw video; (ii) low-to-moderate motion, ruling out high-motion protocols. P-Hybrid is the most deployment-robust choice (neutral on pre-compressed MPEG-4, winning on clean and low-SAC conditions)~(Fig.~\ref{fig:h3}).
\end{enumerate}

\section{Methods}
\label{sec:methods}

\subsection{SAC Definition}
\label{sec:sac_def}

Spatial Artifact Coherence (SAC) is computed per subject-variant cell as follows:
\begin{enumerate}
\item Extract the four-patch Green-channel time series across all 10-second windows.
\item Apply a 0.75--2.5\,Hz bandpass filter.
\item Compute the $4{\times}4$ covariance matrix across patches.
\item
\begin{equation}
\SAC = \frac{\text{mean}(|\text{off-diagonal elements}|)}{\text{mean}(|\text{diagonal elements}|)}
\label{eq:sac}
\end{equation}
\end{enumerate}
SAC is not equivalent to the mean pairwise Pearson correlation across patches: it measures the scale of cross-patch absolute covariance relative to mean patch variance, preserving the absolute coherence magnitude that is lost when each pair is individually normalised.
SAC also differs from frequency-domain spatial coherence functions used in array signal processing, which compute normalised cross-power spectral density in the frequency domain; SAC operates on bandpass-filtered covariance energy in the time domain and yields a scalar boundary condition rather than a spectral estimator.
SAC~$= 0$ indicates spectrally independent patches in the cardiac band; SAC~$= 1$ indicates all patches co-vary identically. The per-cell SAC is the mean across windows within each subject-variant pair. The primary analysis pools MCD FullHDwebcam ($N = 200$) and both UBFC datasets ($N = 80$), yielding 3080 unique cells across 280 subjects.

\subsection{Datasets}

Three public datasets were used: MCD~\cite{b16}, UBFC-rPPG~\cite{b17}, and UBFC-PHYS~\cite{b18}. Detailed specifications are listed in Table~\ref{tab:datasets}.

\begin{table}[!t]
\centering
\caption{Dataset Specifications. Source bitrates measured by ffprobe on 10 files per camera (2026-04-16). All MCD cameras output 640$\times$480; ``FullHDwebcam'' is the device marketing name, not the output resolution.}
\label{tab:datasets}
\setlength{\tabcolsep}{3pt}
\resizebox{\columnwidth}{!}{%
\begin{tabular}{llllrp{1.5cm}}
\toprule
Dataset & Camera & Res. & Source codec & Bitrate & Task \\
\midrule
MCD & FullHDwebcam & 640$\times$480 & MPEG-4 FMP4 & 1346\,kbps & Naturalistic \\
MCD & USBVideo & 640$\times$480 & MPEG-4 FMP4 & 1567\,kbps & Naturalistic \\
MCD & IriunWebcam & 640$\times$480 & MPEG-4 FMP4 & 792\,kbps & Naturalistic \\
UBFC-rPPG & Single cam & 640$\times$480 & Uncompressed & $\sim$218\,000\,kbps & Controlled \\
UBFC-PHYS & Single cam & 1024$\times$1024 & MJPEG & $\sim$245\,000\,kbps & T1/T2/T3 \\

\bottomrule

\end{tabular}
}
\end{table}

\subsection{Degradation Variants}

Eleven degradation variants were applied to all source videos programmatically at parse time via a unified transform module. Each variant simulates a distinct compressed-channel scenario encountered in deployed rPPG systems. Table~\ref{tab:variants} lists all variants with their codec family and per-dataset mean SAC values; SAC computation is defined in Section~\ref{sec:sac_def}.

\begin{table}[!t]
\centering
\caption{Degradation Variants. SAC values are dataset means (Section~\ref{sec:sac_def}). MPEG-4 uses libxvid at the stated target bitrate; H.265/H.264 use libx265/libx264 with CRF\,=\,28 and stated Group-of-Pictures (GOP) size. UBFC column reports UBFC-rPPG means.}
\label{tab:variants}
\setlength{\tabcolsep}{3pt}
\footnotesize
\begin{tabular}{p{2.2cm}lcc}
\toprule
Display name & Codec family & SAC (MCD) & SAC (UBFC) \\
\midrule
None (clean) & --- & 0.13 & 0.20 \\
YUV420 & Chroma & 0.17 & 0.22 \\
JPEG (q50) & JPEG & 0.12 & 0.14 \\
JPEG (q30) & JPEG & 0.10 & 0.11 \\
H.264 intra & H.264 & 0.11 & 0.14 \\
H.264 GOP15 & H.264 & 0.12 & 0.11 \\
H.265 intra & H.265 & 0.14 & 0.14 \\
H.265 GOP15 & H.265 & 0.13 & 0.09 \\
MPEG-4 (85\,kbps) & MPEG-4 & 0.59 & 0.68 \\
MPEG-4 (200\,kbps) & MPEG-4 & 0.59 & 0.67 \\
MPEG-4 (500\,kbps) & MPEG-4 & 0.59 & 0.67 \\
\bottomrule
\end{tabular}
\end{table}

Eight non-MPEG-4 variants cover chroma subsampling, JPEG, and H.264/H.265 intra and inter-frame coding. Three MPEG-4 variants span a 6$\times$ bitrate range (85--500\,kbps) to test whether SAC elevation is amplitude-dependent. YUV420 tests PCA's data-driven de-weighting of degraded chroma channels.

\subsection{Face Detection and Patch Extraction}

Face detection and patch RGB extraction are performed using the MediaPipe face mesh landmark detector. Four patches are extracted per frame: forehead, bilateral upper cheeks, bilateral lower cheeks, and nose--chin region. Frames where detection fails are discarded. Patch RGB signals are extracted at the native video frame rate to preserve spectral fidelity in the cardiac band (0.75--2.5\,Hz), then downsampled to 20\,Hz before evaluation. Face detection failure rates are below 2\% per subject across all cameras.

\subsection{Algorithms Evaluated}
\label{sec:methods_alg}

Thirteen algorithms are evaluated (Table~\ref{tab:algorithms}). Three are established baselines; five are the proposed PatchPCA family; five are diagnostic variants appearing in supplementary material only. CHROM~\cite{b3} is the primary comparison: it is the most widely cited chrominance-projection method and represents the upper bound of published global-projection rPPG. POS~\cite{b4} performs within 0--3\,BPM of CHROM across all conditions, so advantages demonstrated over CHROM are conservative (equal or larger advantages hold over POS and 2SR). The PatchPCA variants are novel algorithms and are not implementations of prior published methods.

\begin{table}[!t]
\centering
\caption{Algorithms Evaluated. Diagnostic variants are not included in the main analysis.}
\label{tab:algorithms}
\setlength{\tabcolsep}{3pt}
\footnotesize
\begin{tabular}{p{2.0cm}p{0.9cm}p{3.8cm}}
\toprule
Label & Role & Key design element \\
\midrule
CHROM & Baseline & Fixed chrominance projection~\cite{b3} \\
POS & Baseline & Plane-orthogonal-to-skin~\cite{b4} \\
2SR & Baseline & Second spectral redundancy~\cite{b5} \\
\midrule
P-Hybrid & Novel & CHROM baseline + PCA on CHROM residuals \\
P-Motion\_Adaptive & Novel & Motion-gated PCA; P-frame specialist \\
P-SpatialTemporal & Novel & Spatial-temporal joint decomposition \\
PatchAvg\_Bandpass & Novel & Coherence-exploiting patch average \\
P-CodecRobust & Novel & MPEG-4-targeted; negative design result \\
\midrule
P-EigGap & Diag. & PCA + eigenvalue gap \\
POS\_RGB & Diag. & POS in linear RGB \\
CHROM\_CR & Diag. & CHROM + chroma smoothing \\
P-PURE\_ENTROPY & Diag. & Entropy selection (fails all) \\
P-IPPC\_RBDrift & Diag. & Compounding correction (zero wins) \\
\bottomrule
\end{tabular}
\end{table}

CHROM and POS degrade monotonically with compression but do not fail catastrophically; their fixed projections cannot be co-opted by a correlated artefact. 2SR is susceptible under 4:2:0 subsampling because compression-induced cross-channel correlation mimics haemodynamic coupling. PatchPCA is explicitly spatial: its performance is governed by SAC. All three baselines are retained to bracket the rPPG paradigm and isolate the spatial coherence mechanism.

P-Hybrid anchors on the CHROM signal extracted from the forehead patch, normalised and bandpass-filtered to the cardiac band: $\hat{s}_\text{c}$. For each of the four patches, the normalised bandpassed Green signal $\tilde{g}_i$ is orthogonally projected against $\hat{s}_\text{c}$ to obtain a residual $r_i = \tilde{g}_i - \langle \tilde{g}_i,\hat{s}_\text{c}\rangle\hat{s}_\text{c}$; PCA is applied to the $T \times 4$ residual matrix and the component most correlated with the mean cross-patch residual is selected as $\hat{s}_\text{res}$. The output is $\hat{s}_\text{c} + 0.5\,\hat{s}_\text{res}$: the CHROM baseline refined by the dominant residual pattern not captured by the physical model. The weight 0.5 was fixed by hand after qualitative inspection; no formal hyperparameter search was conducted.

\subsection{Evaluation Protocol}
\label{sec:eval_protocol}

Heart rate is estimated from 10-second, 20\,Hz segments (200 samples per window), with up to 20 windows per subject selected by face-detection coverage. Pulse traces are bandpass filtered at 0.75--2.5\,Hz and HR is estimated by FFT peak detection. The primary metric is MAE (BPM) between estimated HR and ground-truth PPG HR. Secondary metrics are SNR (dB) in the cardiac band (UBFC datasets only; MCD SNR is unavailable due to naturalistic recording noise) and Pearson $r$ between the extracted rPPG trace and ground-truth contact PPG waveform (UBFC-rPPG only).

The motion index for a 10-second window is the mean per-frame Euclidean magnitude of successive head-rotation changes in yaw and pitch, estimated from MediaPipe face mesh landmarks:
\begin{equation}
m_t = \sqrt{(\Delta\,\mathrm{yaw}_t)^2 + (\Delta\,\mathrm{pitch}_t)^2}
\end{equation}
where $\Delta$ denotes the frame-to-frame difference (units: degrees per frame at 20\,Hz); the window motion index is $\bar{m} = T^{-1}\sum_t m_t$.

Statistical significance is assessed per cell (algorithm $\times$ dataset $\times$ camera $\times$ variant) using the Wilcoxon signed-rank test on paired MAE differences. The delta convention throughout is $\Delta = \text{PCA\,MAE} - \text{CHROM\,MAE}$: negative values indicate PCA wins; positive values indicate CHROM wins. All reported significance levels survive Benjamini-Hochberg FDR correction at $q < 0.05$ across all 904 tests (473 significant tests, zero demoted by FDR).

\begin{table*}[t]
\centering
\caption{$\Delta$MAE (BPM) = PCA MAE $-$ CHROM MAE on MCD ($N = 200$, FullHDwebcam). Negative values indicate PCA wins. Significance: $^{***}p < 0.001$, $^{**}p < 0.01$, $^{*}p < 0.05$. All tests FDR-corrected ($q < 0.05$, 904 tests).}
\label{tab:delta_mcd}
\scriptsize
\begin{tabularx}{\textwidth}{l *{11}{>{\centering\arraybackslash}X}}
\toprule
Algorithm & None & YUV & JPG-50 & JPG-30 & H264i & H264g & H265i & H265g & MP4-85k & MP4-200k & MP4-500k \\
\midrule
PAvgBP   & $-0.49$     & $-0.70^{**}$  & $-1.22^{***}$ & $-1.05^{*}$   & $-0.72$     & $-0.23$     & $-1.20^{***}$ & $-1.56^{***}$ & $+6.36^{***}$ & $+6.48^{***}$ & $+6.50^{***}$ \\
P-SpTmp  & $-0.64$     & $-0.68$       & $-1.06$       & $-0.80$       & $-1.27^{**}$& $-0.55$     & $-1.52^{***}$ & $+0.55^{***}$ & $+0.97^{***}$ & $+1.03^{***}$ & $+0.95^{***}$ \\
P-MotAdp & $-0.23$     & $-0.29$       & $-0.85$       & $-0.75$       & $-0.74$     & $-0.45$     & $-1.05^{*}$   & $-1.32^{***}$ & $+2.93^{***}$ & $+2.93^{***}$ & $+2.91^{***}$ \\
P-Hybrid & $-0.20$     & $-0.10^{*}$   & $-0.41$       & $-0.07$       & $-0.59$     & $-0.37$     & $-0.60$       & $+0.31^{**}$  & $+0.15$       & $+0.20$       & $+0.22$ \\
P-CdcRb  & $-0.03^{*}$ & $-0.01$       & $-0.68$       & $-0.28$       & $-0.95$     & $-0.14^{*}$ & $-0.82$       & $+0.20^{***}$ & $+2.42^{***}$ & $+2.35^{***}$ & $+2.25^{***}$ \\
\bottomrule
\multicolumn{12}{p{\textwidth}}{\scriptsize H264i = H.264 intra, H264g = H.264 GOP15, H265i = H.265 intra, H265g = H.265 GOP15, MP4 = MPEG-4 at stated bitrate, JPG = JPEG. P-Hybrid is the only algorithm with near-zero $\Delta$ on all three MPEG-4 columns (all NS).}
\end{tabularx}
\end{table*}

\begin{table*}[t]
\centering
\caption{$\Delta$MAE (BPM) = PCA MAE $-$ CHROM MAE on UBFC-PHYS ($N = 40$). Negative values indicate PCA wins.}
\label{tab:delta_ubfcphys}
\scriptsize
\begin{tabularx}{\textwidth}{l *{11}{>{\centering\arraybackslash}X}}
\toprule
Algorithm & None & YUV & JPG-50 & JPG-30 & H264i & H264g & H265i & H265g & MP4-85k & MP4-200k & MP4-500k \\
\midrule
PAvgBP   & $+1.48$           & $+2.72^{***}$       & $-0.36$             & $+0.48$             & $+1.63^{**}$        & $+1.13^{**}$        & $-0.54$             & $-1.49$             & $-0.57$             & $+0.15$             & $-0.49$ \\
P-SpTmp  & $-1.49$           & $-0.60$             & $-1.67^{*}$         & $-1.24$             & $-0.64$             & $-0.62$             & $-2.00^{*}$         & $-1.15$             & $-0.84$             & $-0.36$             & $-0.79$ \\
P-MotAdp & $+0.25$           & $+1.53^{*}$         & $-1.42$             & $-1.96$             & $+0.20$             & $-0.13$             & $-0.89$             & $-1.11$             & $-0.74$             & $-0.31$             & $-0.68$ \\
P-Hybrid & $-0.72$           & $-0.89$             & $-1.90^{*}$         & $-1.31$             & $-1.15$             & $-0.94$             & $-1.69^{*}$         & $-1.74$             & $-0.93$             & $-0.42$             & $-0.90$ \\
P-CdcRb  & $-0.78$           & $-0.78$             & $-1.29$             & $-2.04$             & $-0.70$             & $-1.43$             & $-2.73^{***}$       & $-1.05$             & $-0.85$             & $-0.22$             & $-0.65$ \\
\bottomrule
\end{tabularx}
\end{table*}

\begin{table*}[t]
\centering
\caption{$\Delta$MAE (BPM) = PCA MAE $-$ CHROM MAE on UBFC-rPPG ($N = 40$). Negative values indicate PCA wins. MPEG-4 columns are strongly positive for most algorithms (PCA fails on raw-source MPEG-4); P-Hybrid is the only algorithm with systematically negative $\Delta$ on non-MPEG-4 variants.}
\label{tab:delta_ubfcrppg}
\scriptsize
\begin{tabularx}{\textwidth}{l *{11}{>{\centering\arraybackslash}X}}
\toprule
Algorithm & None & YUV & JPG-50 & JPG-30 & H264i & H264g & H265i & H265g & MP4-85k & MP4-200k & MP4-500k \\
\midrule
PAvgBP   & $+1.15$             & $+1.96$             & $+3.05$             & $+4.07$             & $+1.51$             & $-1.81$             & $+1.26$             & $-3.22^{**}$        & $+10.74^{***}$      & $+10.71^{***}$      & $+10.48^{***}$ \\
P-SpTmp  & $+0.84$             & $+2.21^{*}$         & $+2.14$             & $+1.26$             & $+0.48$             & $+1.04$             & $-2.15$             & $+0.76$             & $+11.41^{***}$      & $+11.21^{***}$      & $+11.40^{***}$ \\
P-MotAdp & $+0.58$             & $+2.53$             & $+1.78$             & $+1.29$             & $+1.05$             & $+0.40$             & $+0.68$             & $-0.78$             & $+11.05^{***}$      & $+10.32^{***}$      & $+11.00^{***}$ \\
P-Hybrid & $-3.94$             & $-4.49^{**}$        & $-1.26$             & $-1.12$             & $-2.04$             & $-2.86$             & $-4.01$             & $+0.01$             & $+5.41^{**}$        & $+4.87^{**}$        & $+5.14^{**}$ \\
P-CdcRb  & $+0.33$             & $+0.40$             & $+1.13$             & $+1.08$             & $+0.70$             & $+0.66$             & $-0.05$             & $+0.98$             & $+11.75^{***}$      & $+10.95^{***}$      & $+11.70^{***}$ \\
\bottomrule
\multicolumn{12}{p{\textwidth}}{\scriptsize UBFC-rPPG has $N = 40$, so many non-MPEG-4 results are NS despite consistent direction (underpowered for moderate effect sizes). P-Hybrid's wins ($-3.94$ to $-4.49^{**}$) are the strongest single-algorithm results in the dataset.}
\end{tabularx}
\end{table*}

\section{Results}
\label{sec:results}

\subsection{Spatial Artifact Coherence Determines PCA Viability}

SAC (Section~\ref{sec:sac_def}) is bimodal across the 11 degradation variants, with zero overlap between codec families: non-MPEG-4 variants cluster at SAC 0.10--0.18 while MPEG-4 variants cluster at SAC 0.48--0.59. At the codec-family level, SAC explains 93.8\% of between-variant variance in PCA advantage ($r = +0.969$ across 11 variants). At the subject level within any single variant, the mean SAC--delta correlation is $r = +0.099$, confirming that SAC is a codec-family classifier rather than a subject-level predictor; individual outcomes within a fixed variant are governed by motion, skin tone, and lighting rather than within-condition SAC fluctuations.

\subsection{Algorithm-Level $\Delta$MAE Heatmap}

Key observations: P-Hybrid is the only algorithm with near-zero $\Delta$ on all three MPEG-4 variants in MCD ($+0.15$ to $+0.22$\,BPM, all not significant). P-MotAdp's flagship result is H.265 GOP15 on MCD ($-1.32$\,BPM, statistically significant at $p < 0.001$). P-SpTmp wins on intra codecs (H.264 intra statistically significant at $p < 0.01$, H.265 intra at $p < 0.001$) but fails H.265 GOP15 ($+0.55^{***}$) and all MPEG-4. PAvgBP wins H.265 GOP15 on UBFC-rPPG ($-3.22^{**}$) and JPEG on MCD (statistically significant at $p < 0.001$ and $p < 0.05$). All MPEG-4 columns on UBFC-rPPG are strongly positive ($+5$ to $+12$\,BPM) for all algorithms except P-Hybrid.

\subsection{Algorithm Profiles: MCD}

\begin{figure}[!t]
\centerline{\includegraphics[width=\columnwidth]{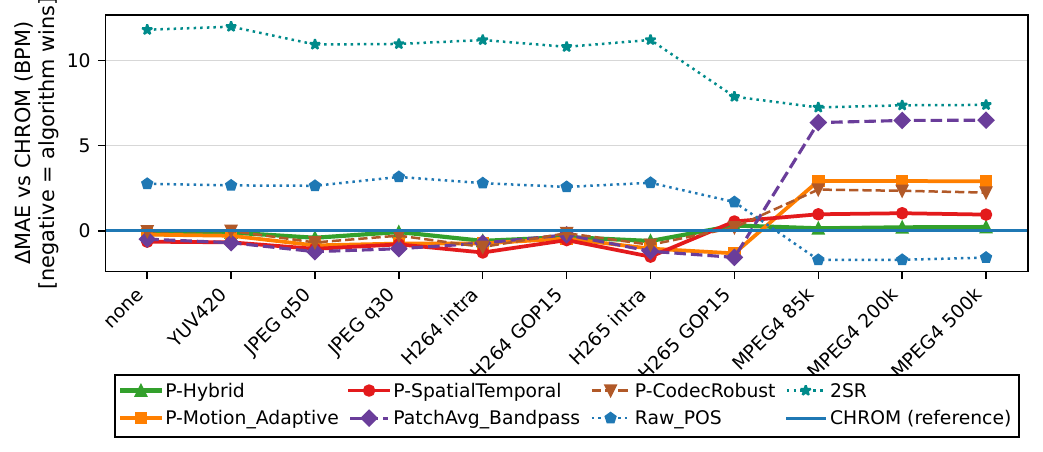}}
\caption{$\Delta$MAE versus CHROM (BPM) for primary algorithms across all 11 variants, MCD. Dotted line separates non-MPEG-4 (left) from MPEG-4 (right). P-MotAdp peaks at H.265 GOP15 ($\Delta = -1.32$\,BPM, $p < 0.001$); PAvgBP degrades most severely under MPEG-4 ($\Delta > +6$\,BPM); P-Hybrid stays near zero on all MPEG-4 variants.}
\label{fig:profiles}
\end{figure}

Fig.~\ref{fig:profiles} shows the delta-MAE profiles for all algorithms across all 11 variants on MCD.

The three MPEG-4 variants occupy the three hardest conditions despite spanning a 6$\times$ bitrate range, confirming the mechanism is spatial structure, not noise amplitude. H.265 GOP15 is the fourth-hardest condition for CHROM yet is the codec regime where P-MotAdp wins most reliably; codec difficulty for CHROM does not equal difficulty for PCA.

\subsection{H3: H.265 GOP15 P-Frame Advantage}

The H3 hypothesis tests whether PCA degrades less than CHROM when switching from H.265 intra-only (spatial DCT only) to H.265 GOP15 (adds inter-frame P-frame prediction). Table~\ref{tab:h3} summarises the interaction effect per dataset and camera group.

\begin{table}[!t]
\centering
\caption{H3 Interaction: CHROM MAE Increase (intra $\rightarrow$ GOP15) Minus Best PCA MAE Increase. Positive interaction = PCA degrades less.}
\label{tab:h3}
\setlength{\tabcolsep}{3pt}
\footnotesize
\begin{tabular}{lrrrr}
\toprule
Dataset / Camera & CHROM $\Delta$ & Best PCA $\Delta$ & Interaction & Algorithm \\
\midrule
MCD pooled        & $+4.21$\,BPM & $+3.93$\,BPM & $+0.28$ & P-MotAdp \\
MCD / IriunWebcam & $+9.93$\,BPM & $+6.51$\,BPM & $\mathbf{+3.41}$ & P-MotAdp \\
UBFC-PHYS         & --         & --         & $+0.95$ & PAvgBP \\
UBFC-rPPG         & --         & --         & $\mathbf{+4.48}$ & PAvgBP \\
\bottomrule
\multicolumn{5}{p{8.5cm}}{\footnotesize H.264 GOP15 does NOT show this effect (all datasets not significant). This specificity to H.265 supports the B-frame diversity hypothesis.}
\end{tabular}
\end{table}

\begin{figure}[!t]
\centerline{\includegraphics[width=\columnwidth]{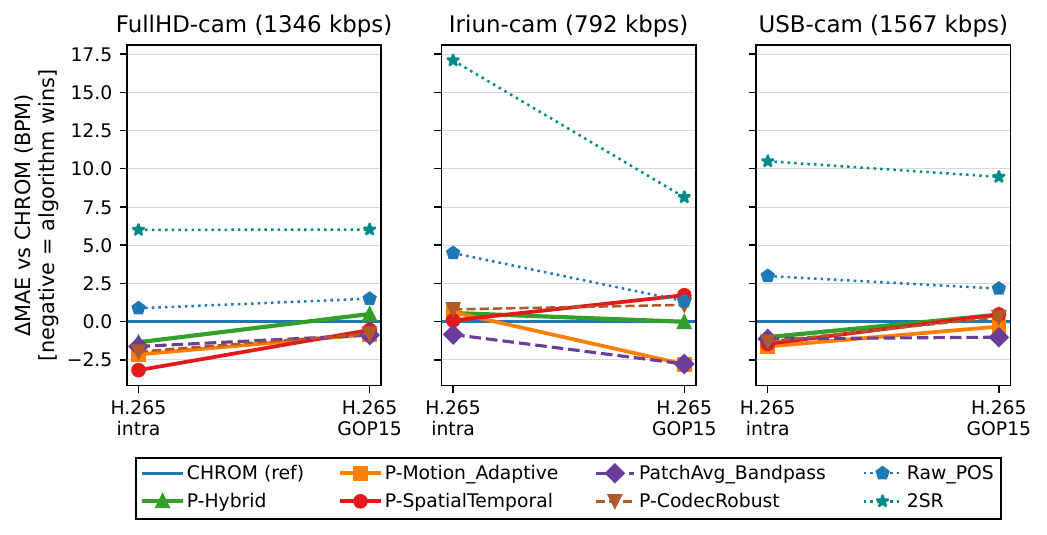}}
\caption{$\Delta$MAE versus CHROM at H.265 intra-only and H.265 GOP15 for three MCD camera groups. The horizontal line at $\Delta = 0$ represents CHROM. P-MotAdp (dashed red) is the only algorithm whose advantage over CHROM improves when moving from intra to inter-frame encoding, confirming spatially diverse P-frame residuals as the mechanism. The effect is largest for IriunWebcam (H3 interaction $= +3.41$\,BPM). H.264 GOP15 does not show this effect, confirming H.265-specific B-frame diversity.}
\label{fig:h3}
\end{figure}

Critically, H.264 GOP15 does not show this effect: CHROM and PCA degrade equally from H.264 intra to H.264 GOP15 on all datasets. This specificity to H.265 is consistent with the hypothesis that H.265 B-frames and hierarchical motion estimation create spatially richer prediction residuals than H.264 P-frames alone. The IriunWebcam result ($+3.41$\,BPM interaction) is the strongest because its lower source bitrate ($\sim$792\,kbps MPEG-4) means additional inter-frame coding creates larger, more spatially diverse residuals (Fig.~\ref{fig:h3}).

\subsection{Source Codec State and MPEG-4 SAC}

All UBFC-rPPG and MCD recordings use 640$\times$480 resolution; resolution cannot explain different MPEG-4 outcomes. The variable is source video compression state. Table~\ref{tab:source_state} compares the three source states.

\begin{table*}[!t]
\centering
\caption{Source Codec State, Resulting MPEG-4 SAC, and PCA Outcome. All videos are 640$\times$480.}
\label{tab:source_state}
\setlength{\tabcolsep}{8pt}
\begin{tabular}{llrll}
\toprule
Dataset & Source codec & MPEG-4 SAC & PCA outcome & Explanation \\
\midrule
UBFC-rPPG & Raw uncompressed & 0.68 & \textbf{FAILS} ($+5$\,BPM, $p<0.01$) & First compression: max DCT coherence \\
MCD & MPEG-4 FMP4 & 0.59 & Neutral ($+0.15$\,BPM, NS) & Re-encode: cascaded, lower coherence \\
UBFC-PHYS & MJPEG near-lossless & 0.48 & Holds ($-0.4$ to $-1.7$\,BPM) & Near-first compress; larger patches \\
\bottomrule
\end{tabular}
\end{table*}

\begin{figure}[!t]
\centerline{\includegraphics[width=\columnwidth]{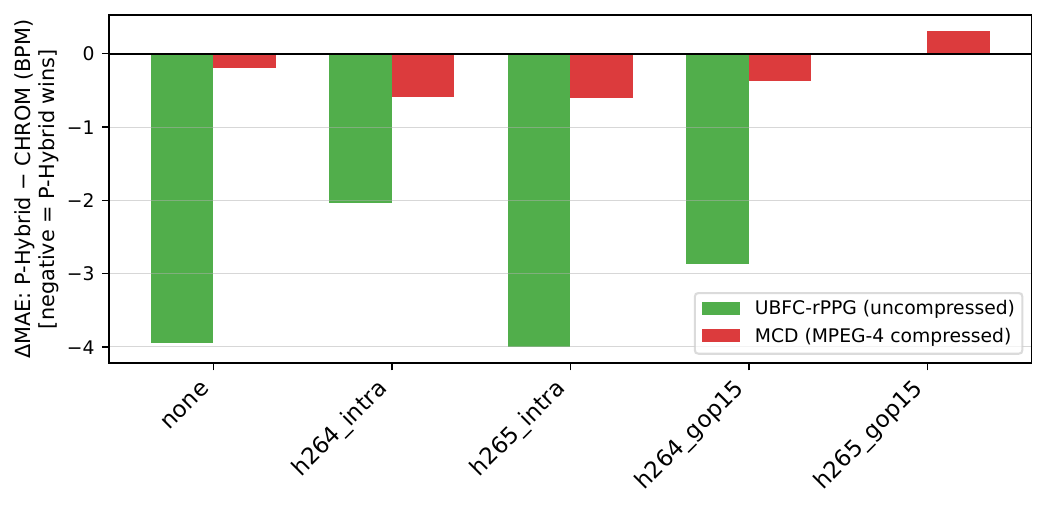}}
\caption{P-Hybrid $\Delta$MAE (BPM) across all 11 variants for UBFC-rPPG (raw source, SAC\,$=0.68$) and MCD (pre-compressed, SAC\,$=0.59$); both at 640$\times$480. Shaded region = MPEG-4. P-Hybrid fails on UBFC-rPPG MPEG-4 ($\Delta \approx +5$\,BPM) but is neutral on MCD ($\Delta \leq 0.22$\,BPM), isolating source codec state as the governing variable.}
\label{fig:source_state}
\end{figure}

For MPEG-4 at 85\,kbps specifically, P-Hybrid achieves $\Delta = +5.41$\,BPM on UBFC-rPPG (PCA fails, $p < 0.01$) versus $\Delta = +0.15$\,BPM on MCD (not significant). CHROM MAE for the same condition is 27.47\,BPM (UBFC-rPPG) and 21.80\,BPM (MCD), so P-Hybrid's absolute performance is also consistent; the failure is relative, driven by CHROM becoming comparatively more robust on raw-source MPEG-4. Fig.~\ref{fig:source_state} visualises this source-state contrast across all 11 variants.

\subsection{Algorithm Verdicts: Unified Summary}

Table~\ref{tab:verdicts} summarises the deployment verdict for each primary algorithm. Two operating boundary conditions for PatchPCA advantage must hold simultaneously: (i) SAC~$< 0.30$, ruling out all MPEG-4 on raw sources; (ii) low-to-moderate motion, ruling out high-motion protocols (UBFC-PHYS T2 at motion index 16.2: zero PCA wins on any codec). Two ablation variants, P-Entropy and P-IRBDrift, produced consistent significant degradation relative to CHROM across all conditions; their behaviour is mechanistically consistent with the SAC framework and informs the design choices of the retained algorithms.

\begin{table*}[!t]
\centering
\caption{Algorithm Performance Summary. Wins and failures refer to statistically significant Wilcoxon results (FDR-corrected).}
\label{tab:verdicts}
\setlength{\tabcolsep}{6pt}
\footnotesize
\resizebox{\textwidth}{!}{%
\begin{tabular}{lllp{4.5cm}}
\toprule
Algorithm & Role & Key wins & Critical failure \\
\midrule
P-Hybrid & \textbf{Flagship (deploy-safe)} & UBFC-rPPG: None ($-3.94$), YUV420 ($-4.49^{**}$), H.265 intra ($-4.01$); UBFC-PHYS: H.265 intra ($-1.69^{*}$) & Raw-source MPEG-4: $+5$\,BPM ($p<0.01$) \\[4pt]
P-MotAdp & \textbf{H.265 GOP15 specialist} & MCD H.265 GOP15: $-1.32^{***}$; IriunWebcam H3: $+3.41$ & MPEG-4; waveform failure on quiet subjects \\[4pt]
P-SpTmp  & \textbf{Intra-codec specialist} & MCD H.265 intra: $-1.52^{***}$; H.264 intra: $-1.27^{**}$; UBFC-PHYS H.265 intra: $-2.00^{*}$ & H.265 GOP15 ($+0.55^{***}$); all MPEG-4 \\[4pt]
PAvgBP   & JPEG / H.265 GOP15 & MCD JPEG (q50): $-1.22^{***}$; JPEG (q30): $-1.05^{*}$; UBFC-rPPG H.265 GOP15: $-3.22^{**}$ & Severe MPEG-4 failure ($>+6$\,BPM) \\[4pt]
P-CdcRb  & Negative design result & UBFC-PHYS H.265 intra only ($-2.73^{***}$) & Fails clean video; MPEG-4 \\
\bottomrule
\end{tabular}
}
\end{table*}

\subsection{PPG Waveform Quality: UBFC-rPPG}

Waveform Pearson $r$ was computed between extracted rPPG traces and ground-truth contact PPG across all UBFC-rPPG variants. MCD and UBFC-PHYS are excluded from waveform analysis (absolute $r \approx \pm 0.01$--$0.03$ for all algorithms in those datasets due to naturalistic noise and task variability). Table~\ref{tab:waveform_r} presents P-Hybrid versus CHROM Pearson $r$ on UBFC-rPPG.

\begin{table}[!t]
\centering
\caption{Waveform Pearson $r$ (rPPG vs GT PPG) on UBFC-rPPG ($N = 40$, all 11 variants). P-Hybrid versus CHROM.}
\label{tab:waveform_r}
\setlength{\tabcolsep}{4pt}
\footnotesize
\resizebox{\columnwidth}{!}{%
\begin{tabular}{lrrrr}
\toprule
Variant & CHROM $r$ & P-Hybrid $r$ & $\Delta r$ & Sig. \\
\midrule
None (clean) & 0.237 & \textbf{0.300} & $+0.063$ & $***$ \\
YUV420       & 0.294 & \textbf{0.330} & $+0.036$ & ns \\
JPEG (q50)   & 0.132 & \textbf{0.170} & $+0.038$ & $**$ \\
JPEG (q30)   & 0.122 & \textbf{0.143} & $+0.021$ & ns \\
H.264 intra  & 0.167 & \textbf{0.245} & $+0.078$ & $***$ \\
H.264 GOP15  & 0.044 & \textbf{0.100} & $+0.055$ & $***$ \\
H.265 intra  & 0.152 & \textbf{0.194} & $+0.042$ & $***$ \\
H.265 GOP15  & 0.019 & 0.011 & $-0.008$ & ns \\
MPEG-4 (85\,kbps)  & 0.007 & \textbf{0.022} & $+0.015$ & $*$ \\
MPEG-4 (200\,kbps) & 0.008 & 0.007 & $-0.001$ & ns \\
MPEG-4 (500\,kbps) & 0.004 & 0.015 & $+0.012$ & ns \\
\bottomrule
\end{tabular}
}
\end{table}

\begin{figure}[!t]
\centerline{\includegraphics[width=\columnwidth]{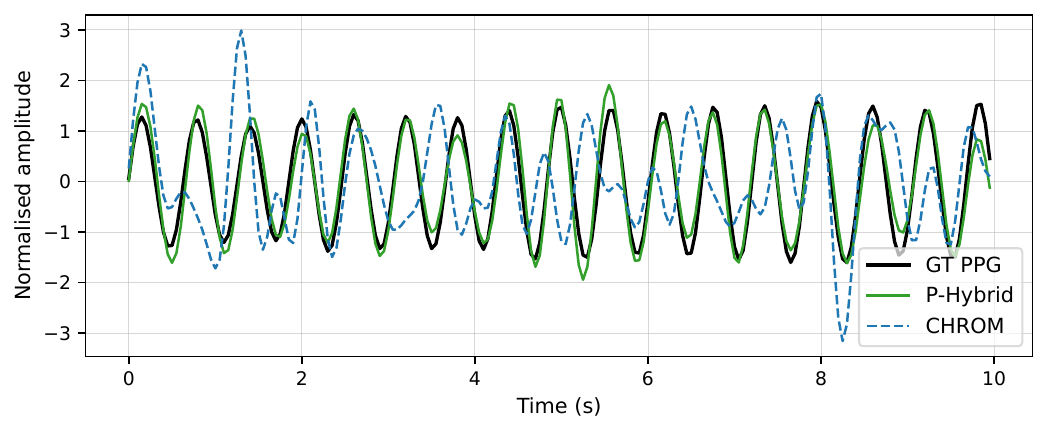}}
\caption{Best-case instantaneous pulse waveform (z-normalised), UBFC-rPPG dataset, None (clean) variant, Subject 42, 10-second window at 660\,s. P-Hybrid (red) achieves Pearson $r = 0.961$ with ground-truth contact PPG (black), tracking individual heartbeat morphology. CHROM (blue dashed) achieves $r = 0.212$. The waveform quality advantage is consistent across all non-MPEG-4 variants (P-Hybrid $r >$ CHROM $r$ at $p < 0.001$ or $p < 0.01$ on 5 of 11 variants; Table~\ref{tab:waveform_r}).}
\label{fig:waveform}
\end{figure}

P-Hybrid achieves statistically significant waveform improvement over CHROM on five variants (None, H.264 intra, H.264 GOP15, H.265 intra, JPEG (q50)) at $p < 0.001$ or $p < 0.01$. Both algorithms collapse to near-zero $r$ under all MPEG-4 variants, confirming that MPEG-4 destroys waveform coherence irrespective of algorithm. The best-case window (Subject 42, 660\,s, None variant) achieves P-Hybrid $r = 0.961$ versus CHROM $r = 0.212$ (Fig.~\ref{fig:waveform}).

P-MotAdp and P-SpTmp show catastrophic waveform degradation: P-MotAdp achieves $\Delta r = -0.252$ ($p < 0.001$) on the None variant, $-0.298$ on YUV420. P-SpTmp shows $\Delta r \approx -0.13$ to $-0.27$ ($p < 0.001$) on all non-MPEG-4 variants. These algorithms' MAE gains under specific codec conditions come at the cost of waveform fidelity on clean or intra-coded video.

\subsection{H2: YUV420 Chroma Subsampling Advantage}

Under 4:2:0 subsampling, PCA's data-driven covariance de-weights degraded Cb/Cr channels while CHROM's fixed projection cannot: P-Hybrid achieves $\Delta = -4.49$\,BPM ($p < 0.01$) on UBFC-rPPG; P-SpTmp achieves $\Delta = -1.91$\,BPM ($p < 0.001$) on MCD. UBFC-PHYS is underpowered ($N = 40$, direction consistent).

\section{Discussion}
\label{sec:discuss}

\subsection{SAC Mechanism and Algorithm Behaviour}

The governing mechanism behind SAC's predictive power is the spatial relationship between MPEG-4's fixed $16{\times}16$ macroblock grid and inter-patch spacing: when macroblock wavelengths span multiple patches, all four patches co-vary identically and PCA's first eigenvector captures the artefact rather than the pulse. When artefacts are spatially diverse (H.265 P-frames, JPEG block noise), PCA isolates the spatially coherent pulse signal. SAC (0.48--0.68 for MPEG-4, 0.10--0.18 for all other families) is consistent across bitrates from 85 to 500\,kbps, confirming that spatial coherence is structural, not amplitude-dependent. CHROM's fixed projection is not susceptible because it cannot be co-opted by a correlated artefact. Source codec state is the primary candidate explanation for SAC level: raw-to-MPEG-4 transcoding (UBFC-rPPG, SAC\,$=0.68$) produces maximum coherence; re-encoding pre-compressed video (MCD, SAC\,$=0.59$) does not recreate first-pass DCT coherence, yielding neutral PCA performance on all MCD MPEG-4 variants.

\textbf{P-Hybrid} anchors on the forehead CHROM signal and refines it with PCA on orthogonal residuals. On clean or low-SAC video, PCA resolves pulse harmonics and fine spatial texture into a waveform refinement (Table~\ref{tab:waveform_r}). On pre-compressed MPEG-4 (MCD, SAC\,$=0.59$), the CHROM baseline degrades gracefully while the orthogonal projection partially de-artefacts the residuals; the residual contribution averages near zero, leaving combined performance at CHROM level. On raw-source MPEG-4 (UBFC-rPPG, SAC\,$=0.68$), first-pass encoding corrupts the forehead CHROM baseline directly; projecting a corrupted baseline contaminates the residuals, PCA amplifies the artefact, and the combination performs worse than CHROM alone ($\Delta = +5.41$\,BPM, $p < 0.01$).

\textbf{P-MotAdp} selectively suppresses H.265 GOP15 P-frame residuals (temporally bursty, spatially variable) via cross-patch variance gating; CHROM accumulates them without discrimination. On quiet UBFC-rPPG subjects the cardiac pulse itself triggers gating, a calibration failure in which the threshold is population-specific rather than universal. Under MPEG-4, high SAC triggers gating on nearly every frame. H.264 GOP15 shows no analogous advantage, confirming H.264 P-frames lack the spatial diversity the mechanism requires.

\textbf{P-SpTmp} separates temporally stationary intra-codec DCT artefacts from the time-oscillating pulse by placing them in different eigenvectors; under H.265 GOP15, P-frame residuals propagate temporally and mimic oscillatory structure, making the stationarity assumption a liability.

\subsection{Boundary Conditions, Deployment Scope, and Limitations}

Two necessary conditions govern PCA advantage: SAC~$< 0.30$ and low-to-moderate motion. Under UBFC-PHYS T2 (mean motion index 16.2), no PCA variant achieves significance on any codec; under T1 (motion index 9.3), wins are significant on intra and H.265 variants. Both conditions must be favourable simultaneously. P-Hybrid is the robust choice for low-SAC conditions; when SAC is elevated (MPEG-4 on raw video), global-projection methods are safer. New codecs (VP9, AV1, H.265 at extreme CRF) should be classified by SAC before algorithm selection.

Several limitations apply. UBFC-rPPG and UBFC-PHYS ($N = 40$ each) are underpowered for moderate effects. The four-patch $4 \times 4$ covariance matrix limits spatial diversity; increasing patch count may shift the SAC boundary. MCD SNR could not be computed due to naturalistic recording conditions. The source-codec-state comparison is observational: MCD and UBFC-rPPG differ in demographics and recording conditions beyond source codec. 10-second windows limit frequency resolution to $\approx$6\,BPM; preliminary 30-second analyses confirm the SAC relationship and directional verdicts. This study evaluates signal-processing rPPG algorithms; the effect of codec-induced SAC on end-to-end deep-learning rPPG methods, which do not explicitly model inter-patch covariance, is left for future work. Most fundamentally, SAC here functions as a binary codec-family classifier (MPEG-4 at SAC~$\sim$0.48--0.59 versus all others at SAC~$< 0.18$, with zero overlap); whether SAC operates continuously across intermediate-SAC codecs (VP9, AV1) remains an open question.

\vspace{-6pt}
\section{Conclusion}
\label{sec:conclusion}
We introduced the PatchPCA algorithm family (four codec-aware rPPG algorithms) and Spatial Artifact Coherence (SAC) as the physical quantity determining boundary conditions under which patch-based PCA outperforms global-projection methods under codec compression. Validated across 280 subjects, three public datasets, 11 codec variants, and 13 algorithms (including CHROM, POS, and 2SR as established baselines), SAC explains 93.8\% of between-variant variance in PCA advantage ($r = +0.969$), with a bimodal distribution producing zero overlap: non-MPEG-4 variants at SAC 0.10--0.18 yield 84--90\% PCA win rates (mean improvement 6.39\,BPM); MPEG-4 at SAC 0.48--0.59 reduces this to 61\% (1.11\,BPM), a 5.8$\times$ collapse. Within subjects, 78\% show the expected pattern ($p < 10^{-22}$, $d_z = 0.73$). MPEG-4's uniquely damaging effect is structural (macroblock-to-patch geometry, not noise amplitude) and is determined by source codec state, not resolution. All reported significance levels survive FDR correction at $q < 0.05$. H.265 GOP15 is the codec regime most favourable to PCA, confirmed by H.264 specificity. We identify P-Hybrid as the most deployment-robust algorithm and empirically establish two necessary operating conditions: SAC~$< 0.30$ and low-to-moderate motion; on raw-source MPEG-4 pipelines (SAC\,$\geq 0.48$), global-projection methods (CHROM) remain the safer choice. SAC enables codec-aware algorithm selection without exhaustive per-codec benchmarking, providing a principled basis for rPPG deployment decisions across the compressed video infrastructure used in clinical remote monitoring.
\vspace{-8pt}


\begin{thebibliography}{00}

\bibitem{b1}
W.~Verkruysse, L.~O.~Svaasand, and J.~S.~Nelson, ``Remote plethysmographic imaging using ambient light,'' \emph{Opt.\ Express}, vol.~16, pp.~21434--21445, 2008.

\bibitem{b2}
M.-Z.~Poh, D.~J.~McDuff, and R.~W.~Picard, ``Non-contact, automated cardiac pulse measurements using video imaging and blind source separation,'' \emph{Opt.\ Express}, vol.~18, pp.~10762--10774, 2010.

\bibitem{b3}
G.~de~Haan and V.~Jeanne, ``Robust pulse rate from chrominance-based rPPG,'' \emph{IEEE Trans.\ Biomed.\ Eng.}, vol.~60, pp.~2878--2886, 2013.

\bibitem{b4}
W.~Wang, A.~C.~den~Brinker, S.~Stuijk, and G.~de~Haan, ``Algorithmic principles of remote PPG,'' \emph{IEEE Trans.\ Biomed.\ Eng.}, vol.~64, pp.~1479--1491, 2017.

\bibitem{b5}
W.~Wang, S.~Stuijk, and G.~de~Haan, ``A novel algorithm for remote photoplethysmography: Spatial subspace rotation,'' \emph{IEEE Trans.\ Biomed.\ Eng.}, vol.~63, pp.~1974--1984, 2016.

\bibitem{b6}
G.~Balakrishnan, F.~Durand, and J.~Guttag, ``Detecting pulse from head motions in video,'' in \emph{Proc.\ CVPR}, 2013, pp.~3430--3437.

\bibitem{b7}
S.~Tulyakov, X.~Alameda-Pineda, E.~Ricci, L.~Wen, J.~S.~Rocha, and N.~Sebe, ``Self-adaptive matrix completion for heart rate estimation from face videos under realistic conditions,'' in \emph{Proc.\ CVPR}, 2016, pp.~2396--2404.

\bibitem{b8}
X.~Li, J.~Chen, G.~Zhao, and M.~Pietik{\"a}inen, ``Remote heart rate measurement from face videos under realistic situations,'' in \emph{Proc.\ CVPR}, 2014, pp.~4264--4271.

\bibitem{b9}
X.~Chen and G.~McDuff, ``DeepPhys: Video-based physiological measurement using convolutional attention networks,'' in \emph{Proc.\ ECCV}, 2018, pp.~349--365.

\bibitem{b10}
X.~Niu, S.~Shan, H.~Han, and X.~Chen, ``RhythmNet: End-to-end heart rate estimation from face via spatial-temporal representation,'' \emph{IEEE Trans.\ Image Process.}, vol.~29, pp.~2409--2423, 2020.

\bibitem{b11}
Z.~Yu, X.~Li, and G.~Zhao, ``Remote photoplethysmograph signal measurement from facial videos using spatio-temporal networks,'' in \emph{Proc.\ BMVC}, 2019.

\bibitem{b12}
A.~Debnath and D.~Kim, ``Remote photoplethysmography: A comprehensive systematic review,'' \emph{Biomed.\ Eng.\ OnLine}, vol.~24, p.~73, 2025.

\bibitem{b13}
D.~J.~McDuff, E.~B.~Blackford, and J.~R.~Estepp, ``The impact of video compression on remote cardiac pulse measurement using imaging photoplethysmography,'' in \emph{Proc.\ CVPRW}, 2017, pp.~1391--1399.

\bibitem{b14}
C.~Zhao, C.-L.~Lin, W.~Chen, and Z.~Li, ``A novel framework for remote photoplethysmography pulse extraction on compressed videos,'' in \emph{Proc.\ CVPRW}, 2018, pp.~1380--1389.

\bibitem{b15}
L.~C.~Parra and P.~Sajda, ``Blind source separation via generalized eigenvalue decomposition,'' \emph{J.\ Mach.\ Learn.\ Res.}, vol.~4, pp.~1261--1269, 2003.

\bibitem{b16}
K.~Egorov, S.~Botman, P.~Blinov \textit{et al.}, ``Gaze into the heart: A multi-view video dataset for rPPG and health biomarkers estimation,'' in \emph{Proc.\ ACM MM}, 2025, pp.~13053--13059.

\bibitem{b17}
S.~Bobbia, R.~Macwan, Y.~Benezeth, A.~Mansouri, and J.~Dubois, ``Unsupervised skin tissue segmentation for remote photoplethysmography,'' \emph{Pattern Recognit.\ Lett.}, vol.~124, pp.~82--90, 2019.

\bibitem{b18}
R.~Sabour, N.~Bendrissou, J.~Benezeth, S.~Bobbia, C.~Guillemot, and Y.~Benezeth, ``UBFC-PHYS: A multimodal database for psychophysiological studies of social stress,'' \emph{IEEE Trans.\ Affect.\ Comput.}, vol.~13, pp.~272--285, 2022.

\bibitem{b19}
E.~M.~Nowara, D.~McDuff, and A.~Veeraraghavan, ``Systematic analysis of video-based pulse measurement from compressed videos,'' \emph{Biomed.\ Opt.\ Express}, vol.~12, pp.~494--511, 2021.

\bibitem{b20}
C.~Zhou, X.~Ye, Y.~Wei \textit{et al.}, ``A comprehensive evaluation of multiple video compression algorithms for preserving BVP signal quality,'' \emph{Biomed.\ Signal Process.\ Control}, vol.~103, 2025.

\bibitem{b21}
B.~Sporrer, N.~Vance, J.~Speth, and P.~Flynn, ``Examining the effects of compression on deep learning remote photoplethysmography,'' in \emph{Proc.\ IS\&T Electron.\ Imag.\ Symp.}, 2024.

\bibitem{b22}
J.~Wang, C.~Shan, Z.~Liu, S.~Zhou, and M.~Shu, ``Physiological information preserving video compression for rPPG,'' \emph{IEEE J.\ Biomed.\ Health Inform.}, vol.~29, pp.~3563--3575, 2025.

\bibitem{b23}
J.~Comas, A.~Ruiz, and F.~Sukno, ``Deep pulse-signal magnification for remote heart rate estimation in compressed videos,'' \emph{Expert Syst.\ Appl.}, 2026, doi:~10.1016/j.eswa.2026.131702.

\end{thebibliography}
\end{document}